%% file: main.tex
\documentclass{article}
\usepackage[preprint]{neurips_2023}
\input{math/math_commands}

\PassOptionsToPackage{sort,square,numbers}{natbib}
\usepackage{url}
\usepackage{booktabs}
\usepackage{amsfonts,mathtools}
\usepackage{nicefrac}
\usepackage{microtype}
\usepackage{wrapfig}
\usepackage{svg}
\usepackage{xspace}
\usepackage{dsfont}
\usepackage{enumitem}
\usepackage{color, colortbl}
\usepackage{pifont}
\usepackage{graphicx}
\usepackage{epstopdf}
\usepackage{pgfplots}
\usepackage{pgfplotstable}
\usepackage{comment}
\usepackage[capitalise]{cleveref}
\usepackage{amssymb}
\usepackage{siunitx}
\usepackage{listings}
\usepackage[utf8]{inputenc}
\usepackage{tabu}
\usepackage{array}
\usepackage{multirow}
\usepackage{diagbox}
\usepackage{amsthm}
\usepackage{placeins}
\usepackage{algorithm,changepage}
\usepackage{float}
\usepackage{balance}
\usepackage{calc}
\usepackage{ifthen}
\usepackage{hyphenat}
\usepackage{fancyhdr}
\usepackage{algorithmic}
\usepackage{eso-pic}
\usepackage{forloop}
\usepackage{subcaption}
\renewcommand{\epsilon}{\varepsilon}

\begin{document}

\title{High-dimensional Bayesian Optimization\\ with Group Testing}
\author{
    Erik Orm Hellsten\thanks{Equal contribution.}\\
    Lund University\\
    \texttt{erik.hellsten@cs.lth.se}\\
    \And
    Carl Hvarfner$^{\ast}$\\
    Lund University\\
    \texttt{carl.hvarfner@cs.lth.se}\\
    \And
    Leonard Papenmeier$^{\ast}$\\
    Lund University\\
    \texttt{leonard.papenmeier@cs.lth.se}\\
    \And
    Luigi Nardi\\
    Lund University, DBtune\\
    \texttt{luigi.nardi@cs.lth.se}\\
}

\maketitle

\newcommand{\data}[0]{\mathcal{D}}
\newcommand{\hps}[0]{\bm{\theta}}
\newcommand{\yx}[0]{y_{\bm{x}}}
\newcommand{\xopt}[0]{\bm{x}^*}
\newcommand{\fopt}[0]{f^*}
\newcommand{\baxus}{\texttt{BAxUS}\xspace}
\newcommand{\hesbo}{\texttt{HeSBO}\xspace}
\newcommand{\turbo}{\texttt{TuRBO}\xspace}
\newcommand{\alebo}{\texttt{Alebo}\xspace}
\newcommand{\rembo}{\texttt{REMBO}\xspace}
\newcommand{\saasbo}{\texttt{SAASBO}\xspace}
\newcommand{\cmaes}{\texttt{CMA-ES}\xspace}
\newcommand{\mctvs}{\texttt{MCTS-VS}\xspace}
\newcommand{\method}{\texttt{GTBO}\xspace}
\newcommand{\xgboost}{\texttt{XGBoost}\xspace}
\newcommand{\fanova}{\texttt{fAnova}\xspace}

\begin{abstract}
Bayesian optimization is an effective method for optimizing expensive-to-evaluate black-box functions.
High-dimensional problems are particularly challenging as the surrogate model of the objective suffers from the curse of dimensionality, which makes accurate modeling difficult.
We propose a group testing approach to identify active variables to facilitate efficient optimization in these domains.
The proposed algorithm, Group Testing Bayesian Optimization (\method), first runs a testing phase where groups of variables are systematically selected and tested on whether they influence the objective.
To that end, we extend the well-established theory of group testing to functions of continuous ranges.
In the second phase, \method guides optimization by placing more importance on the active dimensions.
By exploiting the axis-aligned subspace assumption, \method is competitive against state-of-the-art methods on several synthetic and real-world high-dimensional optimization tasks.
Furthermore, \method aids in the discovery of active parameters in applications, thereby enhancing practitioners' understanding of the problem at hand.
\end{abstract}

\section{Introduction}\label{sec:introduction}
Noisy and expensive-to-evaluate black-box functions occur in many practical optimization tasks, including %
material design~\citep{zhang2020bayesian}, hardware design~\citep{nardi2019practical,ejjeh2022hpvm2fpga}, hyperparameter tuning~\citep{kandasamy2018neural, ru2020interpretable,chen_arXiv18, hvarfner2022joint}, and robotics~\citep{calandra-lion14a, berkenkamp2021safety, mayr2022learning}.
Bayesian Optimization (BO) is an established framework that allows the optimization of such problems in a sample-efficient manner~\citep{shahriari-aistats16a,frazier2018tutorial}.
While BO is a popular approach for black-box optimization problems, its susceptibility to the curse of dimensionality has impaired its applicability to high-dimensional problems, such as in robotics~\citep{calandra-lion14a}, joint neural architecture search and hyperparameter optimization~\citep{bansal2022jahsbench}, drug discovery~\citep{negoescu2011knowledge}, chemical engineering~\citep{burger2020mobile}, and vehicle design~\citep{jones2008large}.

In recent years, efficient approaches have been proposed to tackle the limitations of BO in high dimensions.
Many of these approaches assume the existence of a low-dimensional \textit{active subspace} of the input domain that has a significantly larger impact on the optimization objective than its complement~\citep{wang2016bayesian, letham2020re}.
In many applications, the active subspace is further assumed to be axis-aligned~\citep{nayebi2019framework, eriksson2021high, papenmeier2022increasing, papenmeier2023bounce}, i.e., only a set of all considered variables impact the objective.
The assumption of axis-aligned subspaces is foundational for many successful approaches~\citep{nayebi2019framework,eriksson2021high,song2022monte}.
While it is assumed to hold true for many practical problems, such as in engineering or hyperparameter tuning, few approaches use it to its full extent: %
either they rely on random embeddings that need to be higher-dimensional than the active subspace to have a high probability of containing the optimum~\citep{nayebi2019framework,papenmeier2022increasing}, or they carry out the optimization in the full-dimensional space while disregarding some of the dimensions dynamically~\citep{eriksson2021high, hvarfner2023selfcorrecting, song2022monte}.

Knowing the active dimensions of a problem yields additional insight into the application, allowing the user to decide which problem parameters deserve more attention in the future.
When the active subspace is axis-aligned, finding the active dimensions can be framed as a feature selection problem.
A straightforward approach is first to learn the active dimensions using a dedicated feature selection approach and subsequently optimize over the learned subspace.
We propose to initially find the active dimensions using an information-theoretic approach built around the well-established theory of group testing~\citep{dorfman1943detection}.
Group testing is the problem of finding several active elements within a larger set by iteratively testing groups of elements to distinguish active members.
We develop the theory needed to transition noisy group testing, which otherwise only allows binary observations, to work with evaluations of continuous black-box functions.
This enables group testing in BO and other applications, such as feature selection for regression problems.
The contribution of this work is:
\begin{enumerate}[leftmargin=*,noitemsep]
    \item We extend the theory of group testing to feature importance analysis in a continuous setting tailored towards Gaussian process modeling.
   \item We introduce Group Testing Bayesian Optimization (\method), a novel BO method that finds active variables by testing groups of variables using a mutual information criterion.
    \item We demonstrate that \method frequently outperforms state-of-the-art high-dimensional methods and reliably identifies active dimensions with high probability when the underlying assumptions hold.
\end{enumerate}

\section{Background}\label{sec:background}

\subsection{High-dimensional Bayesian optimization}\label{subsec:high-dimensional-bayesian-optimization}
We aim to find a minimizer $\bm{x}^* \in \argmin_{\bm{x}\in \mathcal{X}} f(\bm{x})$ of the black-box function $f(\bm{x}):\mathcal{X} \rightarrow\mathbb{R}$, over the $D$-dimensional input space $\mathcal{X}=[0, 1]^D$.
We assume that $f$ can only be observed point-wise and that the observation is perturbed by noise, $y(\bm{x}) = f(\bm{x})+\varepsilon_i$ with $\varepsilon_i\sim \mathcal{N}(0, \sigma_n^2)$, where $\sigma_n^2$ is the \emph{noise variance}.
We further assume $f$ to be expensive to evaluate, such that the number of function evaluations is limited.
In this work, we consider problems of high dimensionality $D$, where only $d_e$ dimensions are \emph{active}, and the other ${D-d_e}$ dimensions are \emph{inactive}.
Here, inactive means that the function value changes only marginally along the inactive dimensions in relation to the active dimensions to the extent that satisfactory optimization performance can be obtained by considering the active dimensions alone.
The assumption of active and inactive dimensions is equivalent to assuming an \emph{axis-aligned} active subspace~\citep{eriksson2021high}, i.e., a subspace that can be obtained by removing the inactive dimensions.
We refer the reader to~\cite{frazier2018tutorial} for an in-depth introduction to BO.

\paragraph{Low-dimensional active subspaces and linear embeddings.}
Using linear embeddings is a common approach when optimizing high-dimensional functions that contain a low-dimensional active subspace.
\rembo~\citep{wang2016bayesian} shows that a random embedded subspace with at least the same dimensionality as the active subspace is guaranteed to contain an optimum if the subspace is unbounded.
This inspired the idea to run BO in embedded subspaces.
\alebo~\citep{letham-ba18a} presents a remedy to shortcomings in the search space design of \rembo.
In particular, bounds from the original space are projected into the embedded space, and the kernel in the embedded space is adjusted to preserve distances from the original space.
Other approaches, such as \hesbo~\citep{nayebi2019framework}, and \baxus~\citep{papenmeier2022increasing}, assume the embedded space to be axis-aligned and propose a projection based on the count-sketch algorithm where each dimension in the original space is assigned to exactly one dimension in the embedded space.
\texttt{Bounce}~\citep{papenmeier2023bounce} extends those approaches to combinatorial and mixed spaces with an embedding that allows mixing of various variable types.

\paragraph{High-dimensional Bayesian optimization in the input space.}
Another approach that assumes an axis-aligned active subspace is \saasbo~\citep{eriksson2021high}, which adds a strong sparsity-inducing prior to the hyperparameters of the Gaussian process model.
This makes \saasbo prioritize a low number of active dimensions unless the data strongly suggests otherwise.
It employs a fully Bayesian treatment of the model hyperparameters.
The cubic scaling of the inference procedure lends \saasbo impractical to run for more than a few hundred samples. Another popular approach is to use trust regions~\citep{pedrielli2016g, regis2016trust}.
Instead of reducing the number of dimensions, \turbo~\citep{eriksson2019scalable} optimizes over a hyper-rectangle in input space.
This makes the algorithm more local to counteract the over-exploration exhibited by traditional BO in high dimensions.
Even though \turbo operates in the full input dimensionality and might not scale to arbitrarily high-dimensional problems, it has shown remarkable performance in several applications.

\paragraph{Active subspace learning.}
In this paper, we resolve to a more direct approach, where we learn the active subspace explicitly.
This is frequently denoted by \textit{active subspace learning}.
A common approach is to divide the optimization into two phases.
The first phase consists of selecting points and analyzing the structure to find the subspace.
An optimization phase then follows on the subspace that was identified. The initial phase can also be used alone to gain insights into the problem.
One of the more straightforward approaches is to look for linear trends using methods such as \textit{Principal Component Analysis} (PCA~\citet{ulmasov2016bayesian}) or \textit{Partial Least Squares} (PLS~\citet{bouhlel2016improving}).
\citet{djolonga2013high} use low-rank matrix recovery with directional derivatives with finite differences to find the active subspace.
If gradients are available, the active subspace is spanned by the eigenvectors of the matrix $C:=\int_{\mathcal{X}}\nabla f(x) (\nabla f(x))^Tdx$ with non-zero eigenvalues.
This is used by \citet{constantine2015computing} and \citet{wycoff2021sequential} to show that $C$ can be estimated in closed form for GP regression.
Large parts of the active subspace learning literature yield non-axis-aligned subspaces.
This can be more flexible in certain applications but often provides less intuition about the problem.
We refer to the survey by~\citet{binois2022survey} for a more in-depth introduction to active subspace learning.

\subsection{Group testing}
Group testing~\citep{aldridge2019group} is a methodology for identifying elements with some low-probability characteristic of interest by jointly evaluating groups of elements.
It was initially developed to test for infectious diseases in larger populations but has later been applied in quality control~\citep{cuturi2020noisy}, communications~\citep{wolf1985born}, molecular biology~\citep{balding1996comparative, ngo2000survey}, pattern matching~\citep{macula2004group, clifford2010pattern}, and machine learning~\citep{zhou2014parallel}.

Group testing can be subdivided into two paradigms: \textit{adaptive} and \textit{non-adaptive}.
In adaptive group testing, tests are conducted sequentially, and previous results can influence the selection of subsequent groups, whereas, in the non-adaptive setting, the complete testing strategy is provided up-front.
A second distinction is whether test results are perturbed by evaluation noise.
In the noisy setting, there is a risk that testing a group with active elements would show a negative outcome and vice versa.
Our method presented in Section~\ref{sec:method} can be considered an adaptation of noisy adaptive group testing~\citep{scarlett2018noisy}.

\citet{cuturi2020noisy} present a \emph{Bayesian Sequential Experimental Design} approach for binary outcomes, which at each iteration selects groups that maximize one of two criteria: the first one is the mutual information between the elements' probability of being active, $\bm{\xi}$, in the selected group and the observation.
The second is the area under the marginal encoder's curve~(AUC).
As the distribution over the active group $p(\bm{\xi})$ is a $2^n$-dimensional vector, it quickly becomes impractical to store and update. Consequently, they propose using a \textit{Sequential Monte Carlo}~(SMC) sampler~\citep{del2006sequential}, representing the posterior probabilities by a number of weighted particles.

\section{Group testing for Bayesian optimization}\label{sec:method}
Our proposed method, \method, fully leverages the assumption of axis-aligned active subspaces by explicitly identifying the active dimensions.
This section describes how we adapt the group testing methodology to find active dimensions in as few evaluations as possible.
Subsequently, we use the information to set strong priors for the GP length scales, providing the surrogate model with the knowledge about which features are active.

\paragraph{Noisy adaptive group testing.}
The underlying assumption is that a population of $n$ elements exists, each of which either possesses or lacks a specific characteristic.
We refer to the subset of elements with this characteristic as the active group, considering the elements belonging to this group as active.
We let the random variable $\xi_i$ denote whether the element $i$ is active ($\xi_i=1$), or inactive ($\xi_i=0$), similar to \citet{cuturi2020noisy} who studied binary outcomes.
The state of the whole population can be written as the random vector $\bm{\xi} = \{\xi_1,\ldots,\xi_n\}\in\{0,1\}^n$.
Further, we denote the true state as $\bm{\xi}^*$.

We aim to uncover each element's activeness by performing repeated group tests.
We write $\bm{g}$ as a binary vector $\bm{g}=\{g_1,\ldots,g_n\}\in \{0,1\}^n$, such that $g_i=1$ signifies that element $i$ belongs to the group.
In noisy group testing, the outcome of testing a group is a random event described by the random variable $Y(\bm{g}, \bm{\xi}^*) \in \{0,1\}$.
A common assumption is that the probability distribution of $Y(\bm{g}, \bm{\xi}^*)$ only depends on whether group $\bm{g}$ contains any active elements, i.e., $\bm{g}^\intercal\bm{\xi}^*\ge 1$.
In this case, one can define the sensitivity $$P(Y(\bm{g}, \bm{\xi}^*) =1~|~\bm{g}^\intercal\bm{\xi}^*\ge 1)$$ and specificity $$P(Y(\bm{g}, \bm{\xi}^*)=0~|~\bm{g}^\intercal\bm{\xi}^*= 0)$$ of the test setup.

As we assume the black-box function $f$ to be expensive to evaluate, we select groups $\bm{g}_t$ to learn as much as possible about the distribution $\bm{\xi}$ for a limited number of iterations $t=1\ldots T$.
In other words, we want the posterior probability mass function over $\bm{\xi}$: $$P\left[\bm{\xi}~|~Y(\bm{g}_1)=y_1,\ldots,Y(\bm{g}_T)=y_T\right],$$ to be as informative as possible.

We can identify the active variables by modifying only a few variables in the search space and observing how the objective changes.
Intuitively, if the function value remains approximately constant after perturbing a subset of variables from the default configuration, it suggests that these variables are inactive.
On the contrary, if a specific dimension $i$ is included in multiple subsets and the output changes significantly upon perturbation of those subsets, this suggests that dimension $i$ is highly likely to be active.

Unlike in the traditional group testing problem, where outcomes are binary, we work with continuous, real-valued function observations.
To evaluate how a group of variables affects the objective function, we first evaluate a \emph{default} configuration in the center of the search space, $\bm{x}_\text{def}$, and then vary the variables in the group and study the difference.
We use the group notation $\bm{g}_t\in \{0,1\}^D$ as a binary indicator denoting which variables we change in iteration $t$.
Similarly, we reuse the notation that the random variable $\bm{\xi}$ denotes the active dimensions, and the true state is denoted by $\bm{\xi}^*$.

The new configuration to evaluate is selected as
\begin{align}
    \bm{x}_t &= \bm{x}_{\rm def} \oplus (\bm{g}_t\otimes \bm{U}), \label{eq:point_creation}
\end{align}

where $\bm{U}_{i,j}\sim \mathcal{U}(-0.5,0.5)$, $\oplus$ is element-wise addition, and $\otimes$ is element-wise multiplication.
Note that a point $\bm{x}_t$ is always associated with a group $\bm{g}_t$ that determines along which dimensions $\bm{x}_t$ differs from the default configuration.
For the newly obtained configuration $\bm{x}_t$, we must assess whether $|f(\bm{x_t})-f(\bm{x}_{\rm def})|\gg 0$, which would indicate that the group $\bm{g}_t$ contains active dimensions, i.e., $\bm{g}_t^\intercal \bm{\xi}^* \geq 1$.
However, as we generally do not have access to the true values $f(\bm{x}_{\rm def})$ or $f(\bm{x}_{t})$ due to observation noise, we use an estimate $\hat{f}(\bm{x})$.

Since $f$ can only be observed with Gaussian noise of unknown variance $\sigma_n^2$, there is always a non-zero probability that a high difference in function value occurs between $\bm{x}$ and $\bm{x}_{\rm def}$ even if group $\bm{g}$ contains no active dimensions.
Therefore, we take a probabilistic approach, which relies on two key assumptions:
\begin{enumerate}[leftmargin=*,noitemsep]
    \item $Z_t \coloneqq \hat{f}(\bm{x}_t) - \hat{f}(\bm{x}_{\rm def})\sim \mathcal{N}(0, \sigma_n^2)$ if $\bm{g}_t^\intercal \bm{\xi} = 0$, i.e., function values follow the noise distribution if the group $\bm{g}_t$ contains no active dimensions.
    \item $Z_t \coloneqq \hat{f}(\bm{x}_t) - \hat{f}(\bm{x}_{\rm def})\sim \mathcal{N}(0, \sigma^2)$ if $\bm{g}_t^\intercal\bm{\xi} \geq 1$, i.e., function values are drawn from a zero-mean Gaussian distribution with the function-value variance if the group $\bm{g}_t$ contains active dimensions.
\end{enumerate}

The first assumption follows from the assumption of Gaussian observation noise and an axis-aligned active subspace.
The second assumption follows from a GP prior assumption on $f$, under which $\hat{f}(x_t)$ is normally distributed.
As we are only interested in the change from $f(\bm{x}_\text{def})$, we assume this distribution to have mean zero.

We estimate the noise variance, $\sigma_n^2$, and function-value variance, $\sigma^2$, based on an assumption on the maximum number of active variables, which we set to $\sqrt{D}$. First we evaluate $f$ at the default configuration $\bm{x}_{\text{def}}$. We then split the dimensions into a number of roughly equally sized bins. For each bin, we evaluate $f$ on the default configuration perturbed along the direction of all variables in that bin and compare the result with the default value. We then estimate the function variance as the empirical variance among the $\sqrt{D}$ largest such differences and the noise variance as the empirical variance among the rest. If the assumption holds, there can be no active dimensions in the noise estimate, which is more sensitive to outliers.

Under Assumptions~1 and 2, the distribution of $Z_t$ depends only on whether $\bm{g}_t$ contains active variables.
Given the probability distribution over population states $p(\bm{\xi})$, the probability that $\bm{g}_t$ contains any active elements is
\begin{align}
    p(\bm{g}_t^\intercal \bm{\xi}^*\ge 1) = \sum_{\bm{\xi}\in \{0,1\}^D}\delta_{\bm{g}_t^\intercal \bm{\xi}\geq 1}p(\bm{\xi}). \label{eq:p_gamma_no_smc}
\end{align}

\begin{figure*}[bt!]
    \centering
    \includeinkscape[width=.85\linewidth]{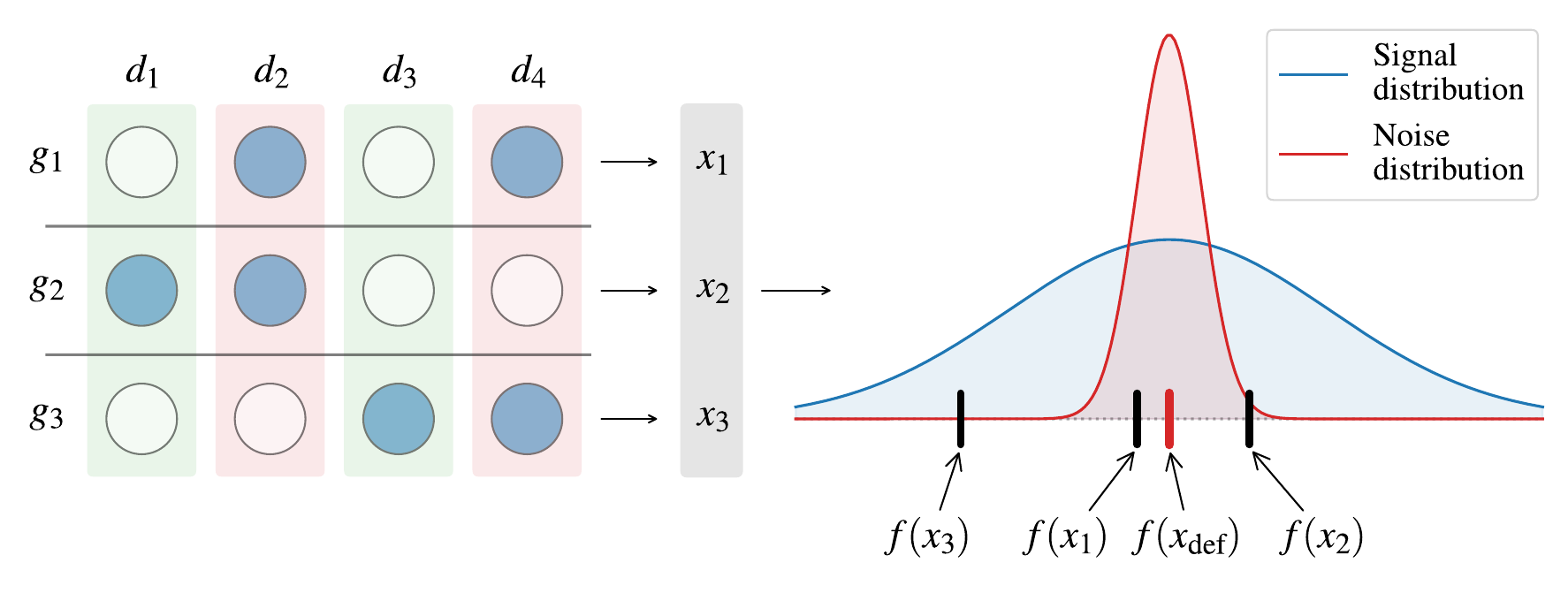}
    \caption{\method assumes an axis-aligned subspace. A point $x_1$ that only varies along inactive dimensions ($d_2$ and $d_4$) obtains a similar function value as the default configuration $(x_{\rm def})$. Points $x_2$ and $x_3$ that vary along active dimensions ($d_1$ and $d_3$) have a higher likelihood under the signal distribution than under the noise distribution.
    }
    \label{fig:gtbo_scheme}
\end{figure*}

We exemplify this in Figure~\ref{fig:gtbo_scheme}.
Here, three groups are tested sequentially, out of which the second and third contain active variables. The three corresponding configurations, $x_1$, $x_2$, and $x_3$, give three function values shown on the right-hand side. As observing $f(x_1)$ is more likely under the noise distribution, $g_1$ has a higher probability of being inactive. Similarly, as $f(x_2)$ and $f(\hat{x}_3)$ are more likely to be observed under the signal distribution, $g_1$ and $g_2$ are more likely to be active.

\paragraph{Estimating the group activeness probability.}
Equation~\eqref{eq:p_gamma_no_smc} requires summing over $2^D$ possible activity states, which, for higher-dimensional functions, becomes prohibitively expensive.
Instead, we use an SMC sampler with $M$ particles $\{\bm{\xi}_1, \ldots, \bm{\xi}_M\}$ and particle weights $\{\omega_1, \ldots, \omega_M\}$.
Each particle $\bm{\xi}_k \in \{0, 1\}^D$ represents a possible ground truth.
We follow the approach presented in \citet{cuturi2020noisy} and use a modified Gibbs kernel for discrete spaces~\citep{liu1996peskun}.
We then estimate the probability $p(\bm{g}_t^\intercal \bm{\xi}^*\ge 1)$ of a group $\bm{g}_t$ to be active by
\begin{align}
    \hat{p}(\bm{g}_t^\intercal \bm{\xi}^*\ge 1) = \sum_{k=1}^M \omega_k \delta_{\bm{g}_t^\intercal\bm{\xi}_k \geq 1}.
\end{align}

\paragraph{Choice of new groups.}
We choose new groups to maximize the information obtained about $\bm{\xi}$ when observing $Z_t$.
This can be achieved by maximizing their \textit{mutual information} (MI).
Under Assumptions~1 and 2, we can write the MI as
\begin{align}
    I(\bm{\xi}, Z_t) &= H(Z_t)-H(Z_t|\bm{\xi}) \label{eq:mi} = H(Z_t)- \sum_{\bm{\xi}\in \{0,1\}^D} p(\bm{\xi})H(Z_t|\bm{\xi})\\
    &= H(Z_t) -  [p(\bm{g}_t^\intercal \bm{\xi}^*\ge 1)H(Z_t | \bm{g}_t^\intercal \bm{\xi} \geq 1)
    + p(\bm{g}_t^\intercal \bm{\xi}^*=0) H(Z_t | \bm{g}_t^\intercal \bm{\xi} =0) ]  \nonumber \\
    &= H(Z_t) - \frac{1}{2}[p(\bm{g}_t^\intercal \bm{\xi}^*=0)\log (2\sigma_n^2\pi e)\nonumber
    +p(\bm{g}_t^\intercal \bm{\xi}^*\ge 1)\log (2\sigma^2 \pi e)].
\end{align}

Since $Z_t$ is modeled as a Gaussian Mixture Model (GMM), its entropy $H(Z_t)$ has no known closed-form expression~\citep{huber2008entropy}, but can be approximated using Monte Carlo:
\begin{align}
    H(Z_t) = \mathbb{E}[- \log Z_t] \approx -\frac{1}{N}\sum_{i=1}^N\log p(z_t^i),
\end{align}
where $z_t^i$ is drawn from $\mathcal{N}(0, \sigma^2)$ with probability $\hat{p}(\bm{g}_t^\intercal \bm{\xi}^*\ge 1)$ and from $\mathcal{N}(0, \sigma_n^2)$ with probability $\hat{p}(\bm{g}_t^\intercal \bm{\xi}^* = 0)$.

\paragraph{Maximizing the mutual information.}
\method optimizes the MI using a multi-start forward-backward algorithm~\citep{russell2010artificial}.
First, several initial groups are generated by sampling from the prior and the posterior over $\bm{\xi}$.
Then, elements are greedily added for each group in a \textit{forward phase} and removed in a subsequent \textit{backward phase}.
In the forward phase, we incrementally include the element that results in the greatest MI increase.
Conversely, in the backward phase, we eliminate the element that contributes the most to MI increase.
Each phase is continued until no further elements are added or removed from the group or a maximum group size is reached.
Finally, the group with the largest MI is returned.

\paragraph{Updating the activeness probability.}
Once we have selected a new group $\bm{g}_t$ and observed the corresponding function value $z_t$, we update our estimate of $\hat{p}(\bm{\xi}_k)$ for each particle $k$:
\begin{align}
    \hat{p}^{t}(\bm{\xi}_k)&\propto \hat{p}^{t-1}(\bm{\xi}_k)p(z_t|\bm{\xi}_k)\\
    &\propto \hat{p}^{t-1}(\bm{\xi}_k)
    \begin{cases}
        p(z_t|\bm{g}_t^\intercal\bm{\xi}_k \geq 1) \,&\text{if}\,  \bm{g}_t^\intercal\bm{\xi}_k \geq 1 \\
        p(z_t|\bm{g}_t^\intercal\bm{\xi}_k = 0 ) \, &\text{if}\,  \bm{g}_t^\intercal\bm{\xi}_k = 0,
    \end{cases}
\end{align}
where $p(z_t|\bm{g}_t^\intercal\bm{\xi}_k = 0)$ and $p(z_t|\bm{g}_t^\intercal\bm{\xi}_k = 1)$ are Gaussian likelihoods.

Assuming that the probabilities of dimensions to be active are independent, the prior probability is given by ${\hat{p}^0(\bm{\xi}_k)= \prod_{i=1}^D q_i^{\bm{\xi}_{k,i}}(1-q_j)^{1-\bm{\xi}_{k,i}}}$ where $q_i$ is the prior probability for the $i$-th dimension to be active.
As we represent the probability distribution $\hat{p}^0(\bm{\xi})$ by a point cloud, any prior distribution can be used to insert prior knowledge.
We use the same SMC sampler as \citet{cuturi2020noisy}.

\paragraph{Batch evaluations.}
Often, several distinct groups can be selected that each have close to optimal mutual information, by simply running the forward-backwards algorithm again excluding already selected groups. When this is possible, we evaluate several groups at once, reducing how often we need to perform the resampling procedure and allowing the user to run several black-box evaluations in parallel. As such, GTBO integrates well with batch BO pipelines with little to no performance degradation.

\paragraph{The \method algorithm.}
With the individual parts defined, we present the complete procedure for \method.
\method iteratively selects and evaluates groups for $T$ iterations or until convergence.
We consider it to have converged when the posterior marginal probability for each variable $\hat{p}^t(\xi_i)$ lies in $[0,C_\text{lower}]\cup [C_\text{upper},1]$, for some convergence thresholds $C_\text{lower}$ and $C_\text{upper}$. More details on the group testing phase can be found in Algorithm~\ref{alg:full} in Appendix~\ref{app:algorithm}.

Subsequently, their marginal posterior distribution decides which variables are selected to be active.
A variable $i$ is considered active if its marginal is larger than some threshold, $\hat{p}^t(\xi_i)\ge \eta$.
Once we have deduced which variables are active, we perform BO using the remaining sample budget.
To strongly focus on the active subspace, we use short lengthscale priors for the active variables and long lengthscale priors for the inactive variables.
We use a Gaussian process (GP) with a Matérn$-\nicefrac{5}{2}$ kernel as the surrogate model and \texttt{qLogNoisyExpectedImprovement} (qLogNEI, \citet{balandat2020botorch}) as the acquisition function.
The BO phase is initialized with data sampled during the feature selection phase.
Several points are sampled throughout the group testing phase that only differ marginally in the active subspace. Such duplicates are removed to facilitate the fitting of the GP.

\section{Computational experiments}\label{sec:computational-experiments}
In this section, we showcase the performance of the proposed methodology, both for finding the relevant dimensions and for the subsequent optimization.
We compare state-of-the-art frameworks for high-dimensional BO on several synthetic and real-life benchmarks.
\method outperforms previous approaches on the tested real-world and synthetic benchmarks.
In Section~\ref{subsec:sensitivity}, we study the sensitivity of \method to external traits of the optimization problem, such as noise-to-signal ratio and the number of active dimensions. The efficiency of the group testing phase is tested against other feature analysis algorithms in Appendix~\ref{sec:comparison_feature_importance}. %
The \method wallclock times are presented in Table~\ref{tab:runtimes} in Appendix~\ref{app:runtimes}.
The code for \method is available at \url{https://github.com/gtboauthors/gtbo}.

\subsection{Experimental setup}
\begin{figure*}[bt!]
     \centering
    \includegraphics[width=.9\textwidth]{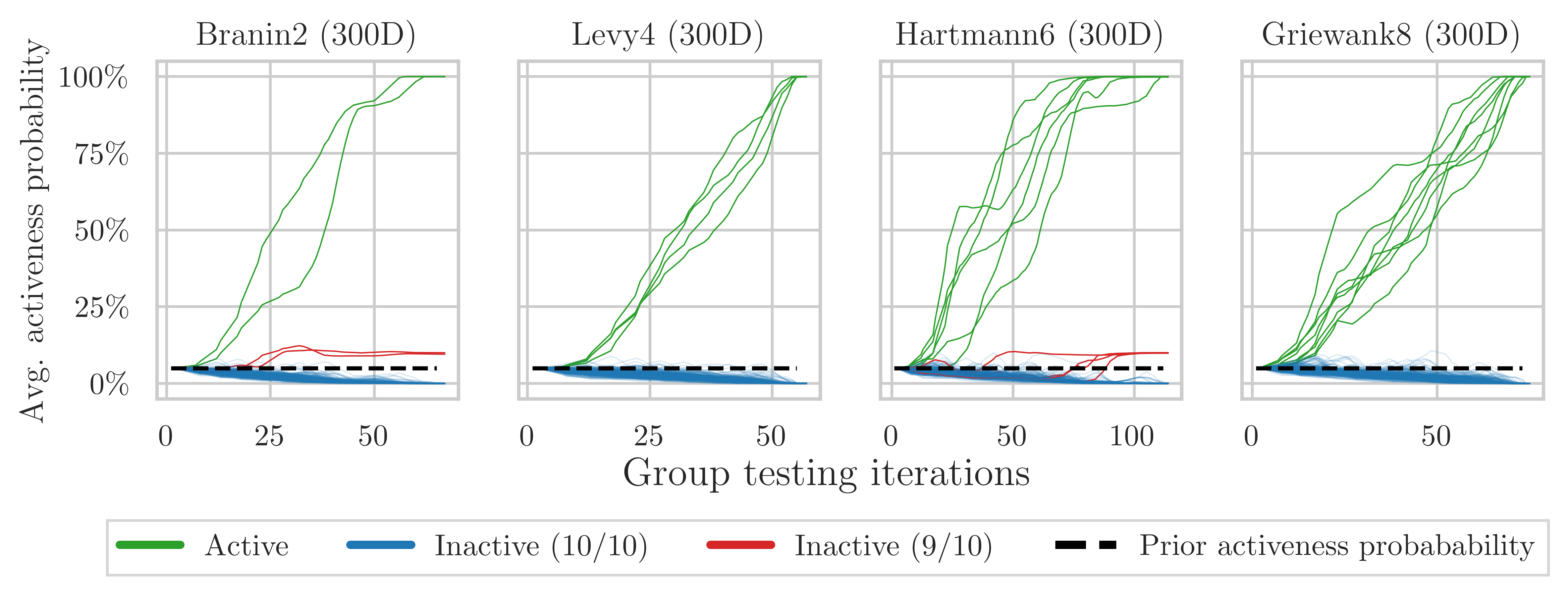}
    \caption{Evolution of the average marginal probability of being active across ten repetitions. Each line represents one dimension, and active dimensions are colored in green and inactive dimensions in blue. In the few cases where \method finds some inactive variables to be active, the lines are emphasized in red. The last iteration marks the end of the \textit{longest} group testing phase across all runs.
    All active dimensions are identified in all runs. 6 out of 1180 inactive dimensions are incorrectly classified as active \textit{once} in ten runs across the benchmarks, implying a false positive rate of slightly above 0.05\%.
    }
    \label{fig:synthetic_gt}
\end{figure*}
We test \method on four synthetic benchmark functions, \texttt{Branin2}, \texttt{Levy} in 4 dimensions, \texttt{Hartmann6}, and \texttt{Griewank} in 8 dimensions, which we extend with inactive ``dummy'' dimensions \citep{wang2016bayesian, eriksson2021high, papenmeier2022increasing} as well as two real-world benchmarks: the 124D soft-constraint version of the \texttt{Mopta08} benchmark \citep{eriksson2021high}, and the 180D \texttt{LassoDNA} benchmark~\citep{vsehic2021lassobench}.
We add significant observation noise for the synthetic benchmarks, but the inactive dimensions are truly inactive.
In contrast, the real-world benchmarks do not exhibit observation noise, but all dimensions have at least a marginal impact on the objective function.
Note that the noisy synthetic benchmarks are considerably more challenging for \method than noiseless problems.

To evaluate the BO performance, we benchmark against \turbo~\citep{eriksson2019scalable} with one and five trust regions, \saasbo~\citep{eriksson2021high}, \cmaes~\citep{hansen1996adapting}, \hesbo~\citep{nayebi2019framework}, and \baxus~\citep{papenmeier2022increasing} using the implementations provided by the respective authors with their settings, unless stated otherwise.
We further compare against random search, i.e., we choose points in the search space $\mathcal{X}$ uniformly at random.

For \cmaes, we use the \texttt{pycma}  implementation~\citep{nikolaus_hansen_2022_6370326}.
For \alebo, we use the \texttt{Ax} implementation~\citep{bakshy2018ae}.
To show the effect of different choices of the target dimensionality $d$, we run \alebo with $d = 10$ and $d = 20$.
We observed that \alebo and \saasbo are constrained by their high runtime and memory consumption.
The available hardware allowed up to 100 evaluations for \saasbo and 300 evaluations for \alebo for each run.
Larger sampling budgets or higher target dimensions for \alebo resulted in out-of-memory errors.
We note that limited scalability was expected for these two methods, whereas the other methods scaled to considerably larger budgets, as required for scalable BO.
We initialize each optimizer with ten initial samples and \baxus with $b = 3$ and $m_D = 1000$ and run ten repeated trials.
Plots show the mean logarithmic regret for synthetic benchmarks and the mean function value for real-world benchmarks. The shaded regions indicate one standard error.

Unless stated otherwise, we run \method with $10000$ particles for the SMC sampler, the prior probability of being active, $q$, is $0.05$, and $3$ initial groups for the forward-backward algorithm. The threshold to be considered active after the group testing phase, $\eta$ is set to $0.5$, and the lower and upper convergence thresholds, $C_\text{lower}$ and $C_\text{upper}$, are $5\times10^3$ and $0.9$, respectively.
We employ a log-normal $\mathcal{LN}(7,1)$ length scale prior to the inactive dimensions for synthetic experiments.
In real-world applications, variables detected inactive can still have a small impact on the function value.
To allow \method to take these variables into consideration, we employ a $\mathcal{LN}(3,1)$ length scale prior to the inactive dimensions for real-world experiments.
Appendix~\ref{app:other_priors} shows the complementary experiments with a $\mathcal{LN}(3,1)$ length scale prior for the synthetic and a $\mathcal{LN}(7,1)$ length scale prior for the real-world experiments. In contrast, we use a $\mathcal{LN}(0,1)$ prior for the active variables, which results in significantly shorter length scales.

The group testing phase on 2x Intel Xeon Gold 6130 machines, using two cores. The subsequent BO part is run on a single Nvidia A40 graphics card supported by Icelake CPUs.

\subsection{Performance of the group testing}\label{ssec:performance_group_testing}
Before studying how well \method works in high-dimensional settings, we study the performance of the group testing procedure.
In Figure~\ref{fig:synthetic_gt}, we show how the average marginal probability of being active evolves over the iterations for the different dimensions. The true active dimensions are plotted in green, and the inactive dimensions are plotted in blue.
For all the problems, \method correctly classifies all active dimensions during all runs within 39-112 iterations.
Across ten runs, \method{} misclassifies 6 out of 1180 inactive variables to be active once each, emphasized in red, for a false positive rate of 0.05\%. How the number of active variables changes over the iterations is shown in Appendix \ref{app:number_active}.

\subsection{Optimization on the relevant variables}\label{subsec:experiments}

\begin{figure*}
    \centering
    \includegraphics[width=\textwidth]{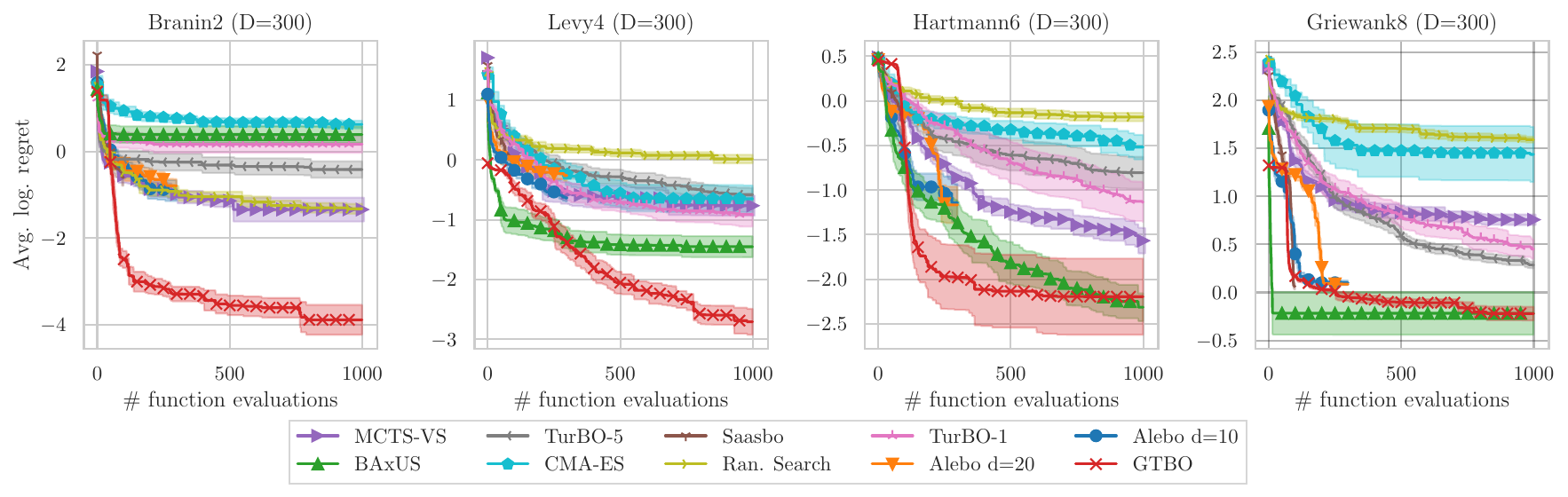}
     \caption{After finding the correct active dimensions, \method optimizes efficiently on synthetic noisy benchmarks (\texttt{Branin2}, \texttt{Levy4}, \texttt{Hartmann6}, and \texttt{Griewank8}). Mean and standard error of average log regret is plotted for each method over 100 iterations.
     }
    \label{fig:gtbo_runs_synth_71}
\end{figure*}

\begin{figure*}
    \centering
    \includegraphics[width=.7\textwidth]{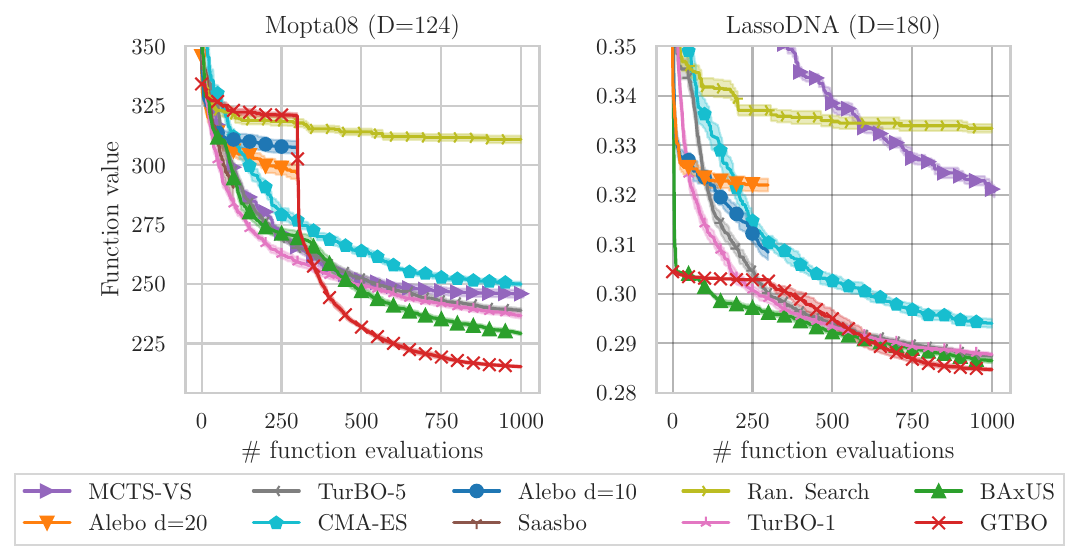}
     \caption{\method outperforms competitors in real-world experiments. Notably, the performance on \texttt{Mopta08} increases by a large amount right after GT at iteration 300, suggesting that the dimensions found during the GT are highly relevant. As the default configuration on \texttt{LassoDNA} performs well, the performance increase after group testing is not as visually apparent.
     }
    \label{fig:gtbo_runs_rw_31}
\end{figure*}

We show that identifying the relevant variables can drastically improve optimization performance.
Figure~\ref{fig:gtbo_runs_rw_31} shows the performance of \method and competitors on the real-world benchmarks, Figure~\ref{fig:gtbo_runs_synth_71} on the synthetic benchmarks.
The results show the incumbent function value for each of the methods, averaged over ten repeated trials.
We plot the true average incumbent function values on the noisy benchmarks without observation noise.

Note that \texttt{Griewank} has its optimum in the center of the search space. To not gain an unfair advantage, we run \method with a non-standard default away from the optimum. However, having the optimum in the center means that all possible projections will contain the optimum, which boosts the projection-based methods \alebo and \baxus.

Figure~\ref{fig:gtbo_runs_rw_31} shows that \method performs well on real-world benchmarks.
Note the drop directly after the group testing phase, indicating that knowing the correct active dimensions drastically speeds up optimization.
In the real-world experiments, especially \texttt{LassoDNA}, the inactive variables still affect the function value.
Appendix~\ref{sec:comparison_feature_importance} shows that \method suffers a slight performance drop when using the stronger priors for the inactive dimensions.
\method uses the center of the search space as a default point for the group testing phase, a decent point in \texttt{LassoDNA}.
Note, however, that \method outperforms the competitors with a faster optimization performance even after starting from a worse solution after the group testing phase.

\subsection{Sensitivity analysis}
\label{subsec:sensitivity}
\begin{figure}[t]
    \centering
    \includegraphics[width=.9\linewidth]{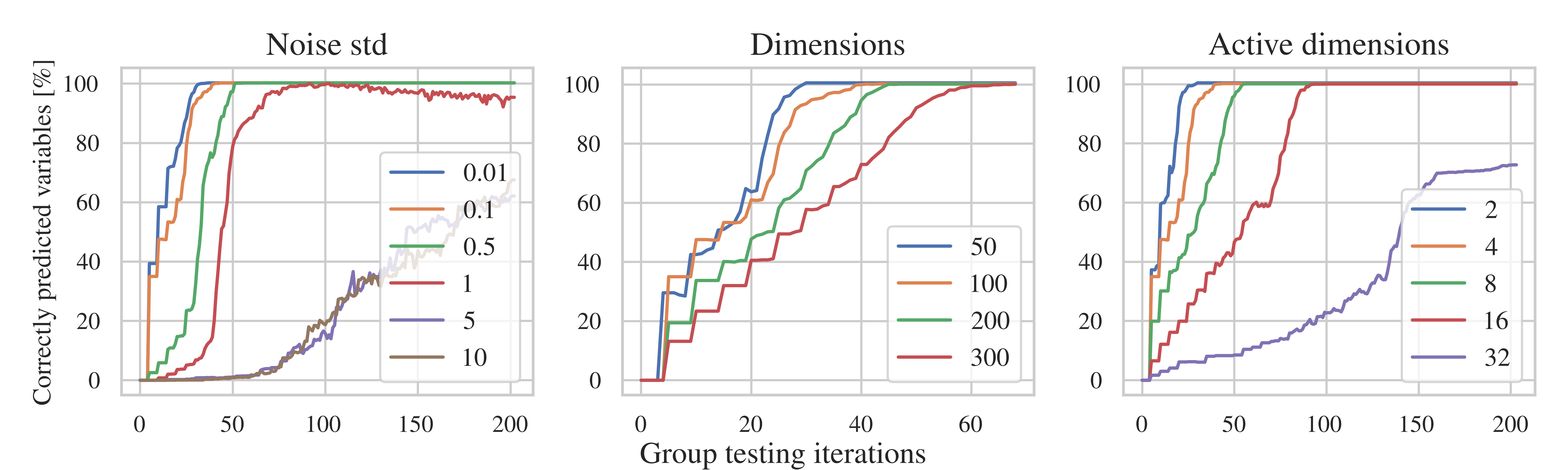}
     \caption{Sensitivity analysis for \method. The average percentage of correctly classified variables is displayed for an increasing number of group testing iterations. The percentage is ablated for (left) various levels of \textit{output noise}, (middle) number of \textit{total dimensions}, and (right) number of \textit{effective dimensions}. Each legend shows the configurations of the respective parameter.}
    \label{fig:sens}
\end{figure}

We explore the sensitivity of \method to the output noise and problem size by running it on the \texttt{Levy4} synthetic benchmark extended to 100 dimensions, with a noise standard deviation of 0.1, and varying the properties of interest.
In Figure~\ref{fig:sens}, we show how the percentage of correctly predicted variables evolves with the number of tests $t$ for different functional properties.
Correctly classified is defined here as having a probability of less than 1\% if inactive or above 90\% if active.
\method shows to be robust with respect to lower noise levels but breaks down when the noise grows too large. As expected, higher function dimensionality and number of active dimensions increases the time until convergence.
Note that the signal and noise variance estimates build on the assumption that there are a maximum of $\sqrt{D}$ active dimensions, which does not hold in the case with 32 active dimensions.%

\section{Discussion}\label{sec:discussion}
Optimizing expensive-to-evaluate high-dimensional black-box functions is a challenge for applications in industry and academia.
We propose \method, a novel BO method that explicitly exploits the structure of a sparse axis-aligned subspace to reduce the complexity of an application in high dimensions.
\method is inspired by the field of group testing in which one aims to find infected individuals by conducting pooled tests but adapts it to the field of Bayesian optimization.

\method quickly detects active and inactive variables and shows robust optimization performance in synthetic and real-world settings.
Furthermore, an important by-product is that users learn what dimensions of their applications are relevant and, consequently, learn something fundamental about their application. Since \method allows for user priors on the activeness of dimensions, we will explore the potential for increasing sample efficiency by including application-specific beliefs.

\section*{Acknowledgements}
Erik Orm Hellsten, Carl Hvarfner, Leonard Papenmeier, and Luigi Nardi were partially supported by the Wallenberg AI, Autonomous Systems and Software Program (WASP) funded by the Knut and Alice Wallenberg Foundation.
Luigi Nardi was partially supported by the Wallenberg Launch Pad (WALP) grant Dnr 2021.0348.
The computations were enabled by resources provided by the National Academic Infrastructure for Supercomputing in Sweden (NAISS) at the Chalmers Centre for Computational Science and Engineering (C3SE) and National Supercomputer Centre at Linköping University, partially funded by the Swedish Research Council through grant agreement no. 2022-06725.
\clearpage
\pagebreak
\FloatBarrier
\bibliography{bibliography/local,bibliography/lib,bibliography/proc,bibliography/strings}
\bibliographystyle{abbrvnat}
\clearpage
\pagebreak
\FloatBarrier
\appendix
\onecolumn
\section{The \method Algorithm}
\label{app:algorithm}
This section describes the group testing phase of the \method algorithm in additional detail.
\FloatBarrier
\begin{algorithm}[H]
\begin{algorithmic}
\REQUIRE black-box function~$f:\mathcal{X}\xrightarrow{}\mathbb{R}$, number of default configuration evaluations~$n_\text{def}$,  number of group tests~$T$, number of particles~$M$, prior distribution~$p^0(\bm{\xi})$
\ENSURE estimate of active dimensions $\bm{\gamma}$, posterior distribution $\hat{p}^T$
\FOR{$j\in \{1\ldots n_\text{def}\}$}
\STATE $y_\text{def}^{(i)} = y(\bm{x}_\text{def})$
\ENDFOR
\STATE $\hat{f}(\bm{x}_\text{def}) \gets \frac{1}{n_\text{def}}
\sum_{i=1}^{n_\text{def}} y_\text{def}^{(i)}$
\STATE
\STATE split the dimensions into $3\lfloor\sqrt{D}\rfloor$ bins $B$
\FOR{$j\in \{1\ldots|B|\}$}
\STATE $y_\text{bin}^{(j)} = y(\bm{x}_\text{def} \oplus \alpha b_j)$ \hfill $\triangleright \,\,$ a random perturbation along the dimensions in bin $b_j$
\ENDFOR
\STATE sort (asc.) $y_\text{bin}^{(j)}$
\STATE $\hat{\sigma}_n^2 \gets \texttt{var} (y_\text{bin}^{(1)},\ldots,y_\text{bin}^{(2\lfloor\sqrt{D}\rfloor)})$
\STATE $\hat{\sigma}^2 \gets \texttt{var} (y_\text{bin}^{(2\lfloor\sqrt{D}\rfloor + 1)},\ldots,y_\text{bin}^{(3\lfloor\sqrt{D}\rfloor)})$
\STATE
\STATE $\bm{\xi}_1, \ldots, \bm{\xi}_M \sim \hat{p}^0(\bm{\xi})$
\STATE $\omega_1, \ldots, \omega_M \gets \frac{1}{M}$ \hfill $\triangleright \,\,$ initial particle weights
\FORALL{$t \in \{1, \ldots, T\}$}
    \STATE $\bm{g}^* \gets $ \texttt{maximize\_mi}$( \bm{\xi}_1, \ldots, \bm{\xi}_M)$ \hfill $\triangleright \,\,$ find a group that maximizes MI
    \STATE $\bm{x}_t \gets $ create using Eq.~\eqref{eq:point_creation} and $\bm{g}^*$
    \STATE $z_t \gets f(\bm{x}_t)+\epsilon - \hat{f}(\bm{x}_\text{def})$
    \STATE $(\bm{\xi}_i, \omega_i)_{i\in [M]}\gets \texttt{resample}(\bm{z}_t,(\bm{\xi}_i, \omega_i)_{i\in [M]})$
    \STATE $\hat{p}^t \gets$ \texttt{marginal}$((\bm{\xi}_i, \omega_i)_{i\in [M]})$ \hfill $\triangleright \,\,$ get marginals
\ENDFOR
\STATE $\bm{\gamma} \gets (\delta_{p^T(\bm{\xi}_1) \geq \eta}, \ldots, \delta_{p^T(\bm{\xi}_D) \geq \eta})$ \hfill $\triangleright \,\,$ check which dimensions are active
\end{algorithmic}
\caption{Group testing phase}
\label{alg:full}
\end{algorithm}

\section{Comparison with feature importance algorithms}
\label{sec:comparison_feature_importance}
We compare the performance of the group testing phase with the established feature importance analysis methods \xgboost~\citep{chen2016xgboost} and \fanova~\citep{hutter2014efficient}.
Since \fanova's stability degrades with increasing dimensionality, we run these methods on the 100-dimensional version of the synthetic benchmarks: \texttt{Branin2} (noise std 0.5), \texttt{Griewank8} (noise std 0.5), \texttt{Levy4} (noise std 0.1), and \texttt{Hartmann6} (noise std 0.01).

\begin{figure}[]
     \centering
     \begin{subfigure}[b]{0.48\textwidth}
         \centering
          \includeinkscape[width=\textwidth]{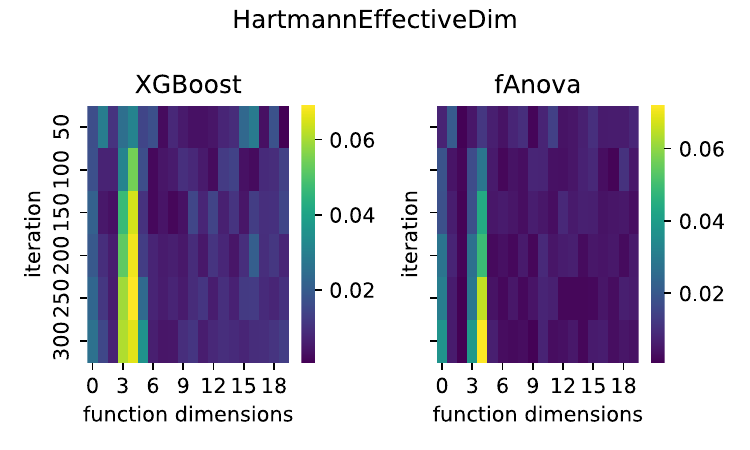}
          \caption{\texttt{Hartmann6}}
          \label{fig:feat_imp_hm6}
     \end{subfigure}
     \hfill
     \begin{subfigure}[b]{0.48\textwidth}
         \centering
          \includeinkscape[width=\textwidth]{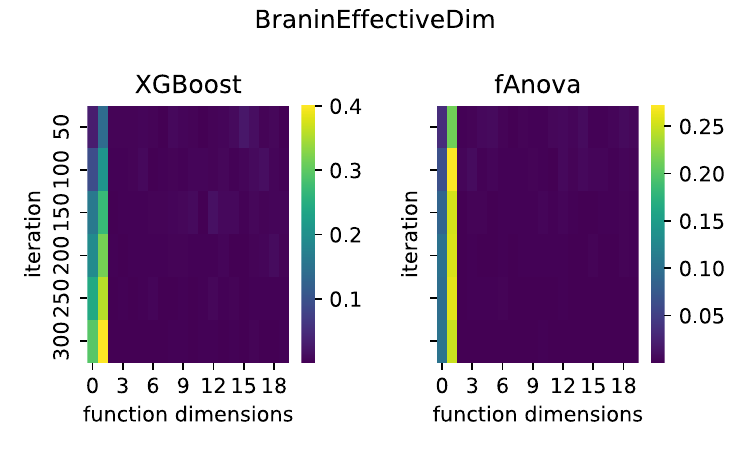}
          \caption{\texttt{Branin2}}
          \label{fig:feat_imp_brn2}
     \end{subfigure}
     \begin{subfigure}[b]{0.48\textwidth}
         \centering
          \includeinkscape[width=\textwidth]{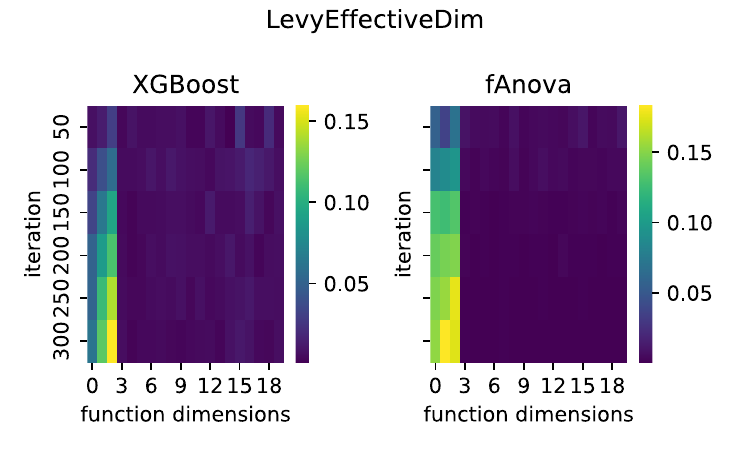}
          \caption{\texttt{Levy4}}
          \label{fig:feat_imp_levy4}
     \end{subfigure}
     \hfill
     \begin{subfigure}[b]{0.48\textwidth}
         \centering
          \includeinkscape[width=\textwidth]{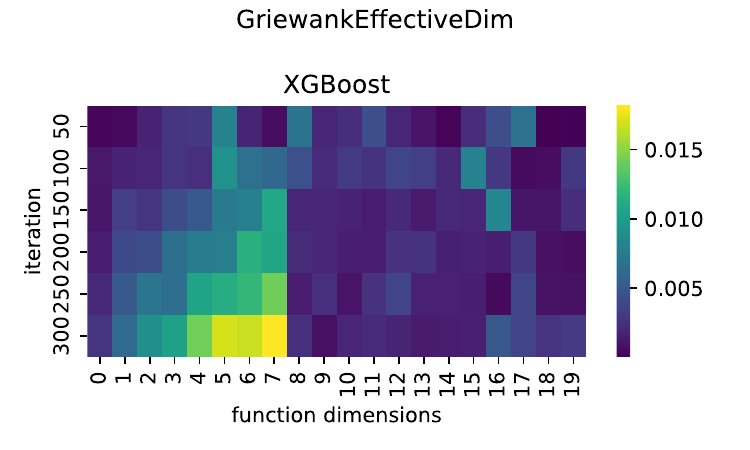}
          \caption{\texttt{Griewank8}}
          \label{fig:feat_imp_gw8}
     \end{subfigure}
     \caption{\xgboost and \fanova feature importance analysis on the 100-dimensional version of different synthetic benchmarks, averaged over 20 repetitions. Only the first 20 dimensions are shown. }
\end{figure}

Figure~\ref{fig:feat_imp_hm6} shows the results of \fanova and \xgboost on the 100-dimensional version of \texttt{Hartmann6} with added output noise.
In accordance with our results, both methods flag the third dimension as not more important than the added input dimensions (dimensions 7-100 with no impact on the function value).
Additionally, \fanova also seems to ``switch off'' the second dimension.

On \texttt{Branin2} (Fig.~\ref{fig:feat_imp_brn2}), both methods detect the correct dimensions (the first and second dimensions). Furthermore, all other dimensions have zero importance, and the methods find the correct partitioning earlier than for \texttt{Hartmann6}. Similarly to \texttt{Hartmann6}, both methods fail to detect an active dimension (the fourth dimension).

\begin{figure}[]
    \centering
    \includegraphics[width=\textwidth]{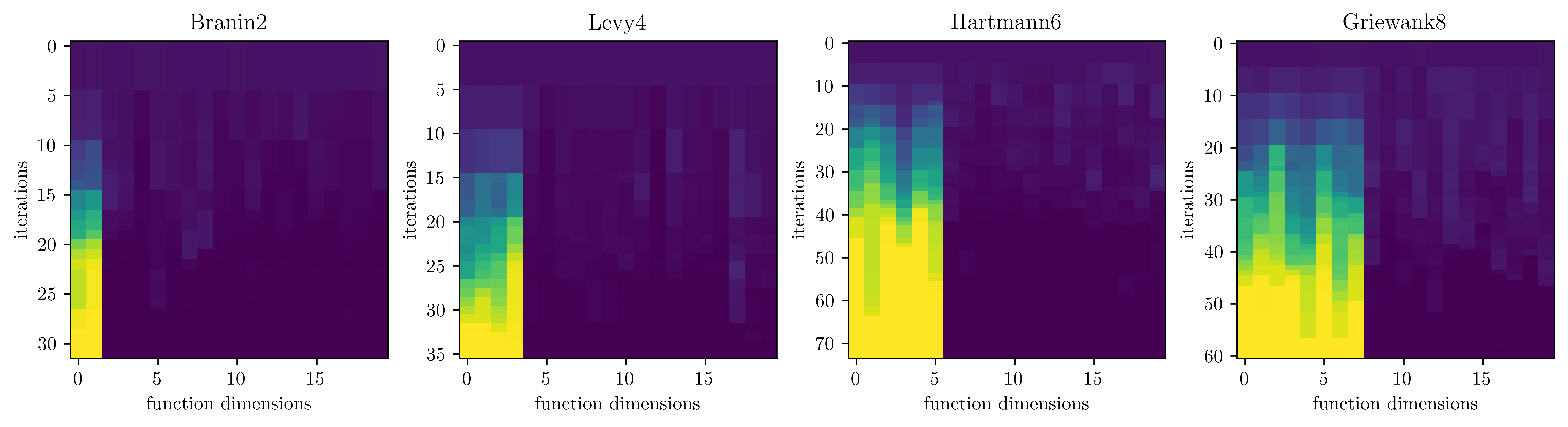}
    \caption{\method-marginals of the first 20 dimensions., averaged over 10 repetitions. If the group testing phase ends early, the last marginals are repeated to match the length of the longest group testing phase.}
    \label{fig:feat_gtbo_marg}
\end{figure}

On \texttt{Griewank8}, \fanova does not terminate gracefully. Therefore, we only discuss XGBoost for \texttt{Griewank8}. After 300 iterations, XGBoost only detects six dimensions reliably as active. The other two dimensions are determined to be not more important than the added dimensions. The marginals found by \method are shown in Fig.~\ref{fig:feat_gtbo_marg}. Compared conventional feature importance analysis methods, \method detects all active dimensions with high probability.

\section{Number of active variables}
\label{app:number_active}
Here, we show the average number of active dimensions throughout the group testing phase. Given that the acceptance threshold of 0.5 is much higher than the initial probability of acceptance of 0.05, dimensions once considered active are rarely later considered inactive again, resulting in a close to monotonically increasing number of active dimensions.
\begin{figure}[]
    \centering
    \includeinkscape[width=\textwidth]{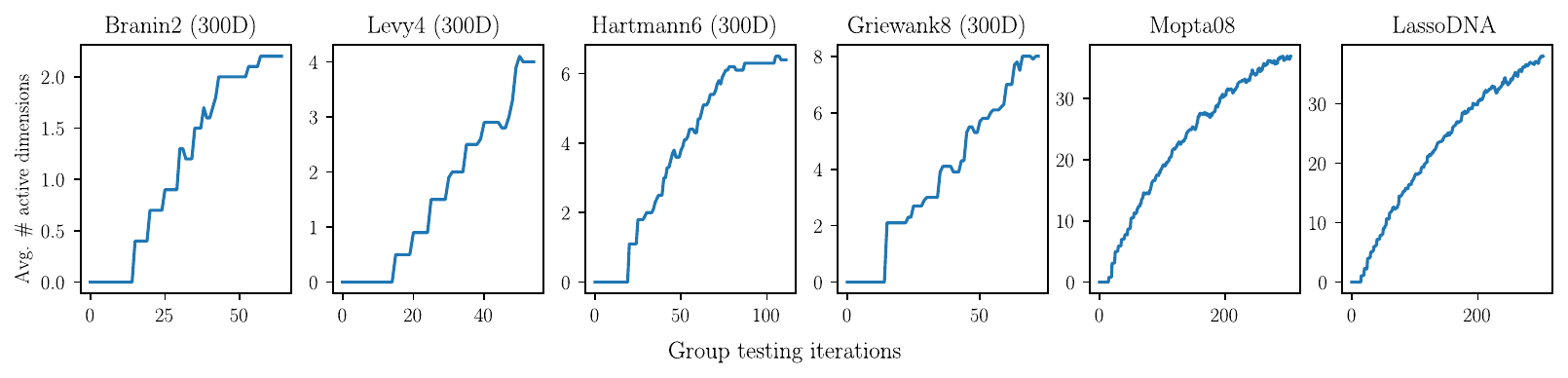}
    \caption{Evolution average number of active variables during the group testing phase (10 repetitions). The synthetic benchmarks find the correct number of active variables, whereas the real-world benchmarks find a significantly higher number.}
    \label{fig:avg_active}
\end{figure}

\section{Complementary Prior Experiments}\label{app:other_priors}

In addition to the lengthscale priors used in Section~\ref{subsec:experiments}, we run the real-world experiments with the complementary priors, i.e., $\mathcal{LN}(7,1)$ length scale priors for the real-world and $\mathcal{LN}(3,1)$ length scale priors for the synthetic experiments.

\begin{figure*}[h]
    \centering
    \includegraphics[width=\textwidth]{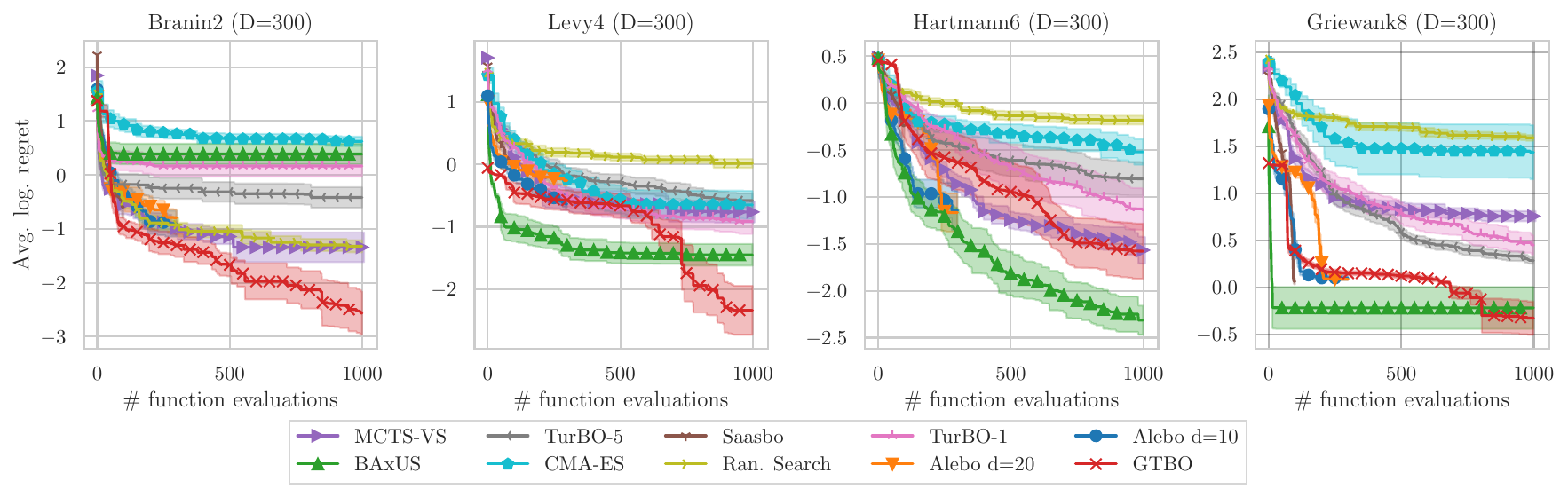}
     \caption{Endowed with a weaker $\mathcal{LN}(3,1)$ length scale prior, \method shows convincing yet slower optimization performance on the noisy synthetic benchmarks (\texttt{Branin2}, \texttt{Levy4}, \texttt{Hartmann6}, and \texttt{Griewank8}).}
    \label{fig:gtbo_runs_synth_31}
\end{figure*}

\begin{figure*}[h]
    \centering
    \includegraphics[width=.8\textwidth]{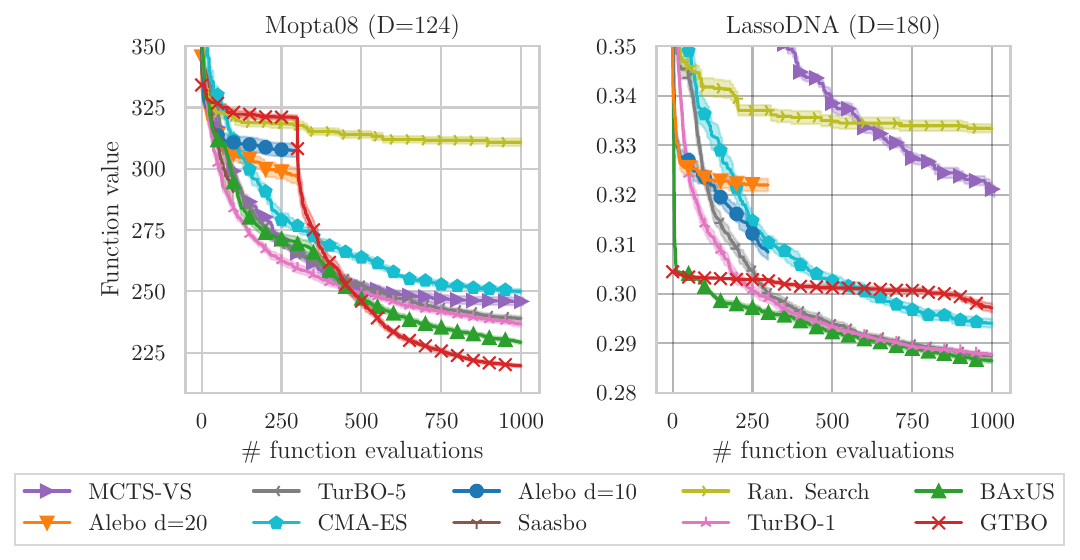}
     \caption{A stronger $\mathcal{LN}(7,1)$ length scale prior still allows \method to achieve state-of-the-art performance on \texttt{Mopta08} and robust performance on \texttt{LassoDNA}.}
    \label{fig:gtbo_runs_rw_71}
\end{figure*}

Figure~\ref{fig:gtbo_runs_synth_31} shows \method's performance on the synthetic benchmarks when equipped with a weaker $\mathcal{LN}(3,1)$ length scale prior.
The weaker prior ``switches on'' dimensions that are found to be inactive during the group-testing phase and slows down optimization performance.
Contrary to this, the stronger $\mathcal{LN}(7,1)$ prior for the off-dimensions for real-world benchmarks (Figure~\ref{fig:gtbo_runs_rw_71}) inhibits \method in its ability to optimize ``off-dimensions''.
We suspect that \texttt{LassoDNA} benefits more from optimizing those dimensions that \texttt{Mopta08}, hence the stronger performance degradation on \texttt{LassoDNA}.

\section{Run times}
\label{app:runtimes}
In this section, we show the average run of \method. Note that the SMC resampling is a part of the group testing phase, and as such, the total algorithm time is the GT time plus the BO time.
We do batch evaluations in the GT phase as described in Section 3, with a maximum of five groups tested before resampling. This significantly reduces the SMC resample time.

\begin{table}[H]
    \centering
    \begin{tabular}{|l|lll|}
            \hline Benchmark & GT time [h] & SMC resample time [h] & BO time [h] \\\hline
            \texttt{Branin2} (300D) & 1.94 & 0.583 & 9.31 \\
            \texttt{Levy4} (300D) & 2.33 & 0.603 & 10.8 \\
            \texttt{Hartmann6} (300D) & 3.29 & 1.14 & 14.1 \\
            \texttt{Griewank8} (300D) & 2.41 & 0.846 & 9.08 \\
            \texttt{Mopta08} (124D) & 5.70 & 4.74 & 7.95 \\
            \texttt{LassoDNA} (180D) & 8.92 & 7.06 & 10.2 \\\hline
    \end{tabular}
    \caption{Average \method runtimes. Group testing time is on the same order of magnitude as the time allocated towards BO. For BO, the $\mathcal{O}(D^2)$ complexity of Quasi-Newton-based acquisition function optimization dominates the runtime.}
    \label{tab:runtimes}
\end{table}
\end{document}

%% file: math/math_commands.tex
\usepackage{amsmath,amsfonts,bm}

\def\1{\bm{1}}

\DeclareMathAlphabet{\mathsfit}{\encodingdefault}{\sfdefault}{m}{sl}
\SetMathAlphabet{\mathsfit}{bold}{\encodingdefault}{\sfdefault}{bx}{n}

\DeclareMathOperator*{\argmin}{arg\,min}